\documentclass[10pt,twocolumn,letterpaper]{article}

\usepackage{iccv}
\usepackage{times}
\usepackage{epsfig}
\usepackage{graphicx}
\usepackage{amsmath}
\usepackage{amssymb}
\usepackage{caption}
\usepackage{subcaption}

\usepackage[pagebackref=true,breaklinks=true,letterpaper=true,colorlinks,bookmarks=false]{hyperref}

\iccvfinalcopy 

\ificcvfinal\fi

\begin{document}

\newcommand\xc{20}
\setlength{\unitlength}{1mm}
\AddToShipoutPicture{%
    \AtTextUpperLeft{
        \put(\numexpr\xc+155\relax,20){
            Originally Submitted to ICCV-21
            }
    }
}

\title{Super-Prompting: Utilizing Model-Independent Contextual Data \\ to Reduce Data Annotation Required in Visual Commonsense Tasks}

\author{
Navid Rezaei and Marek Z. Reformat\\
University of Alberta\\
Edmonton, T6G 1H9, Canada\\
{\tt\small \{nrezaeis,marek.reformat\}@ualberta.ca}
}

\maketitle
\ificcvfinal\thispagestyle{empty}\fi

\begin{abstract}
Pre-trained language models have shown excellent results in few-shot learning scenarios using in-context learning. 
Although it is impressive, the size of language models can be prohibitive to make them usable in on-device applications, 
such as sensors or smartphones. With smaller language models, task-specific data annotation is needed to fine-tune 
the language model for a specific purpose. However, data annotation can have a substantial financial and time burden 
for small research groups, startups, and even companies.

In this paper, we analyze different prompt-based fine-tuning techniques 
to improve results on both language and multimodal causal transformer models. 
To evaluate our results, we use a dataset focusing on visual commonsense reasoning in time.

Our results show that by simple model-agnostic prompt-based fine-tuning, 
comparable results can be reached by only using ~35\%-40\% of 
the fine-tuning training dataset. The proposed approaches result 
in significant time and financial savings.

As the proposed methods make minimal architectural assumptions, 
other researchers can use the results in their transformer 
models with minimal adaptations. We plan to release 
the source code freely to make it easier for the community 
to use and contribute to our work.
\end{abstract}

\section{Introduction}

Human annotation is time-consuming and is also a financial burden for research groups, startups, and companies. To put it in context, 
almost \$240,000 has been spent on the annotation of the Visual Commonsense Reasoning in Time (VisualCOMET) dataset 
and this figure only includes the payment to crowd-workers from Amazon Mechanical Turk \cite{Park2020VisualCOMETRA}. 
The real financial burden can be much higher when including the time value of the staff involved in the annotation process.

Although large pre-trained language models, such as GPT-3 transformer \cite{brown2020language}, 
are impressive at multi-task few-shot learning, their huge size can be prohibitive 
for different scenarios, including on-device applications. Fine-tuning still plays an important role in achieving the state of the art, 
even with a relatively smaller model. As an example, the two current leading models\footnote{https://super.gluebenchmark.com/leaderboard} (better than human baseline) on SuperGLUE task \cite{wang2019superglue} are fine-tuned variants of T5 \cite{2020t5} and DeBERTa \cite{he2020deberta} language models, while GPT-3 is at 14th place.

Our goal is to devise a model-independent process that could improve results based on fine-tuning with much less annotated training data. 

\section{Related Work}
Several recent works have focused on improving fine-tuning methods in language models, 
such as \cite{Howard2018UniversalLM}, \cite{Dodge2020FineTuningPL}, \cite{Lee2020MixoutER}, and \cite{Zhang2020RevisitingFB}. 
The focus has been put mostly on optimization and regularization, but not on using less data for fine-tuning. The results 
from those studies are complementary to our work.

Some previous efforts have been put on prompt-based fine-tuning to improve classification or regression tasks in 
natural language processing (NLP). \cite{schick2020exploiting} and \cite{schick2020its} convert textual inputs into 
cloze-style questions with a task description. \cite{gao2020making} studies smaller language models for few-shot learning 
capability by using automatically-generated prompts for fine-tuning and by incorporating demonstrations into context.

On another topic, a group of recent research studies, including \cite{Houlsby2019ParameterEfficientTL}, \cite{Pfeiffer2020AdapterHubAF} 
and \cite{li2021prefixtuning}, aim at task-dependent added parameters to adapt models to different tasks. 
This way, one does not need to re-train a complete model to fine-tune it to a specific task but only needs to re-train a fraction of parameters.

There is a recent body of work that utilizes inherent knowledge of language models combined with fine-tuning on 
specialized large-scale training datasets to infer different commonsense and causal scenarios. \cite{bosselut-etal-2019-comet} 
uses generative language models to expand on ATOMIC \cite{Sap2019ATOMICAA} and ConceptNet \cite{Speer2017ConceptNet5A} 
commonsense knowledge graphs. \cite{Hwang2020COMETATOMIC2O} introduces an updated knowledge graph similar to ATOMIC and 
uses BART \cite{lewis-etal-2020-bart} encoder-decoder model to generate new knowledge. \cite{mostafazadeh-etal-2020-glucose} 
uses generative language models to expand on an introduced knowledge base of causal mini-story explanations.

Given the success of prompt-based fine-tuning and in-context learning in classification and regression tasks, 
we are motivated to assess similar principles in the context of commonsense generation using generative language models, 
which are fine-tuned on a commonsense knowledge graph.

\section{Dataset}
For this paper, we have selected a multi-modal commonsense knowledge graph for fine-tuning. 
The Visual Commonsense Reasoning in Time (VisualCOMET) dataset \cite{Park2020VisualCOMETRA} 
consists of $1.4$ million commonsense inferences over 59,356 images and 139,377 specific events at present. 
The dataset has human-annotated inferences regarding three different aspects:
the intention of the person mentioned, the possible events that could happen next, and 
the possible preceding events. The inferences are made based on a single image. The annotators
have access to short clips before and after the event, which are not part of the dataset. 
Each image is also annotated with event and place descriptions.
There is a total amount of 1,465,704 commonsense inferences.

The images are sourced from the VCR dataset \cite{Zellers2019FromRT}. The images usually have a complex visual scene 
with multiple people and activities present. This dataset includes automatically-detected object bounding boxes and people 
are annotated with numerical tags.

\section{Method}
In this work, we focus on using generative language models and analyze how prompt-based fine-tuning 
and in-context learning could help to reduce the size of the data required for fine-tuning training.

\begin{figure}
   \centering
   \begin{subfigure}[b]{0.3\textwidth}
       \centering
       \includegraphics[width=0.9\textwidth]{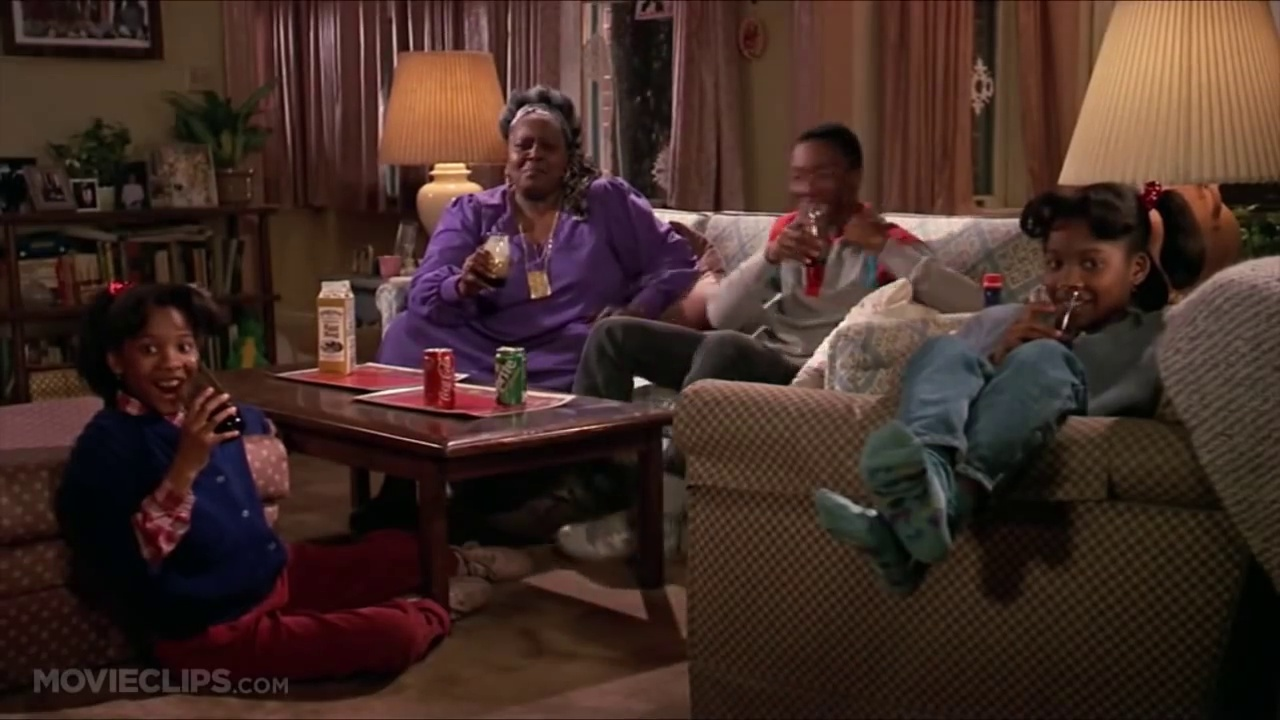}
       \caption{\textbf{Event:} Person-4 is sitting on the couch with her legs over the arm.\\
         	    \textbf{Place:} In a living room\\
         	    \textbf{Annotated Intent Inferences:} \\1) be comfortable \\2) get cozy\\
         	    \textbf{A Predicted Intent:} Show boredom\\
	      	    \textbf{Missing context:} Happy facial expression}
       \label{fig:1a}
   \end{subfigure}
   \hfill
   \begin{subfigure}[b]{0.3\textwidth}
       \centering
       \includegraphics[width=0.9\textwidth]{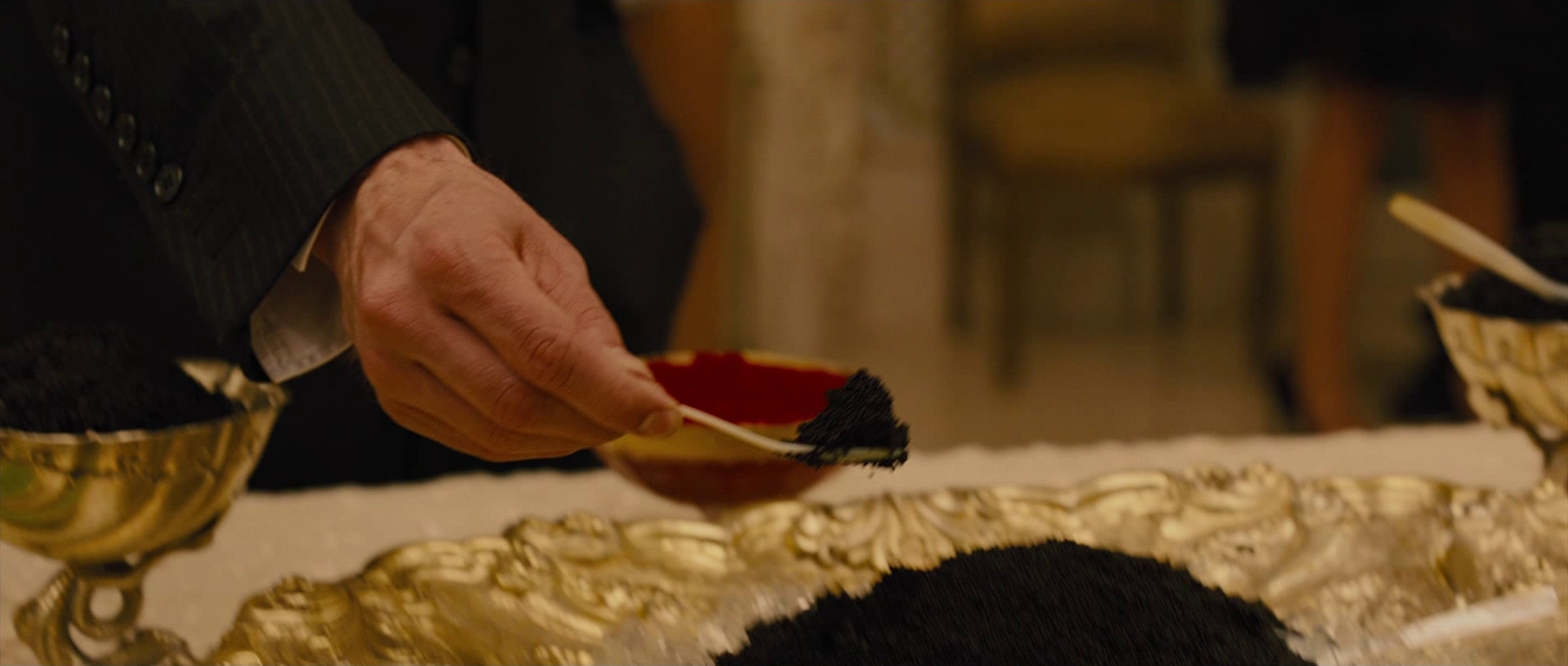}
       \caption{\textbf{Event:} Person-2 is taking a tiny spoon and scooping up a heap of caviar.\\
         	    \textbf{Place:} In a dining room\\
         	    \textbf{Annotated Intent Inferences:} \\1) live luxuriously \\2) enjoy a delicious treat\\
         	    \textbf{A Predicted Intent:} keep everything neat\\
	      	    \textbf{Missing context:} Caviar is a luxury edible.}
       \label{fig:1b}
   \end{subfigure}
   \hfill
   \begin{subfigure}[b]{0.3\textwidth}
       \centering
       \includegraphics[width=0.9\textwidth]{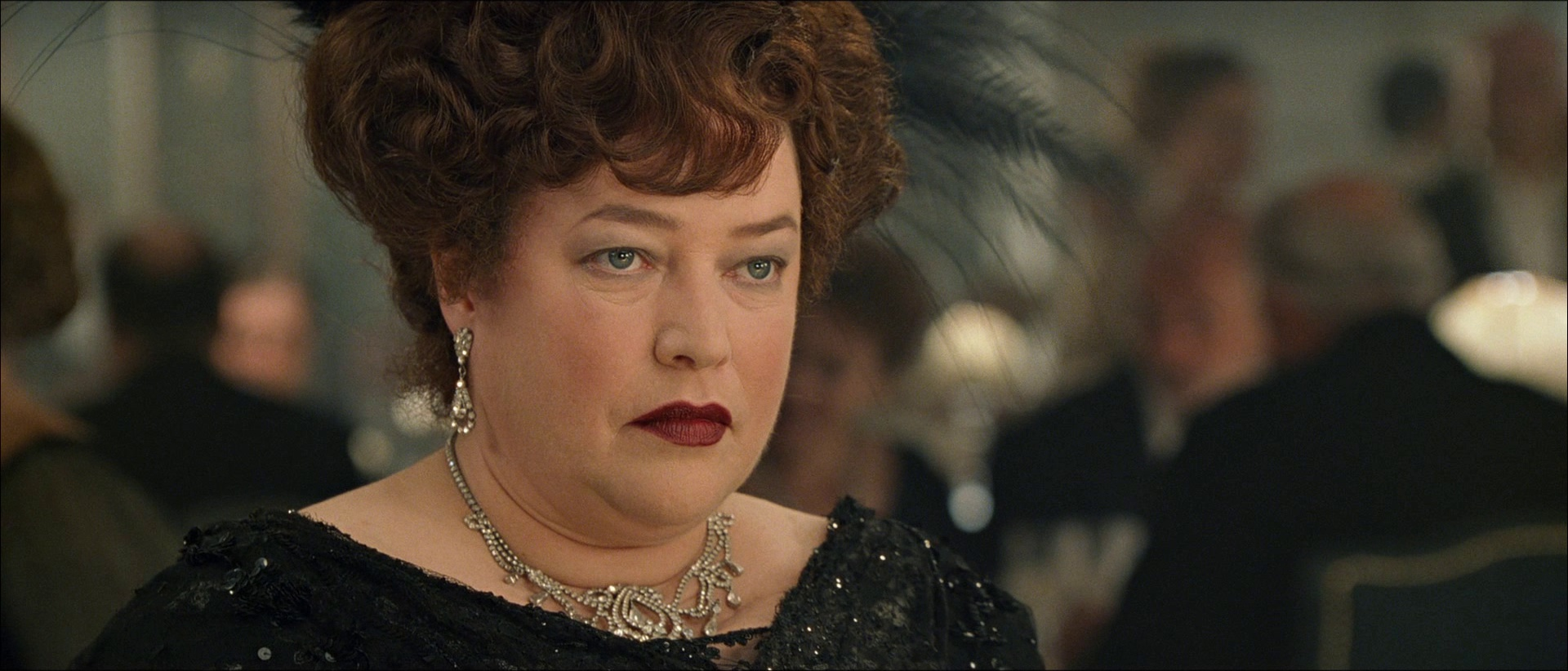}
       \caption{\textbf{Event:} Person-1 is sitting down staring at someone angrily.\\
         	    \textbf{Place:} In a dining room\\
         	    \textbf{Annotated After Inferences:} \\1) choose not to express her anger verbally \\2) look down in front of her \\3) slam the table \\4) walk out of the restaurant\\
         	    \textbf{A Predicted After:} Fight\\
	      	    \textbf{Missing context:} Objects in a dining room and its location.}
       \label{fig:1c}
   \end{subfigure}
      \caption{Predictions based on the fine-tuned language model introduced in \protect\cite{Park2020VisualCOMETRA}. 
               Each example shows a piece of missing contextual information that could be utilized.}
      \label{fig:1}
\end{figure}

As seen in Fig.~\ref{fig:1}, there are several scenarios where extra context could help lead 
the generative language model to a correct answer, but lack of correct understanding about the scene 
and the event text can result in incorrect results. Extra human annotations, focused on these shortcomings, 
could improve the results, but that comes with extra time and money expenditure.

We propose using the underutilized context already present in text and image, then transforming 
them to a form that is usable by most transformer models, which is a sequence. We analyze if 
this kind of addition helps the language model achieve better results in the case of limited annotated data available.

Assuming the added context text is represented with $c$ and its tokenized version with $\{c\}$, 
we can represent the context with $\{c\}= \{w_1^c, w_2^c, ... w_q^c\}$, where $w^c_i$ represents 
each token created from tokenization of the context $c$. This context is merged with tokenized 
versions of event and place, which are represented as: $\{e\}= \{w_1^e, w_2^e, ... w_n^e\}$ and 
$\{p\}= \{w_1^p, w_2^p, ... w_m^p\}$, respectively. Using the merged versions of event and place texts with context, 
the updated sequence-to-sequence loss can be written as:

\begin{align}
    \mathcal{L} = & - \sum_{i=1}^{n}{\log P(w_{i}^{e} | w_{<i}^{e},  v))} - \sum_{i=1}^{m}{\log P(w_{i}^{p} | w_{<i}^{p}, e, v))} \nonumber\\ - & \sum_{i=1}^{q}{\log P(w_{i}^{c} | w_{<i}^{c},  p, e, v))} \nonumber\\ - & \sum_{i=1}^l{\log P(w_{hi}^r | w_{h<i}^r, c, p, e, v)}
\end{align}
where $v$ represents visual features, including overall images and person-specific boxes, 
$r$ represents inference prompts, which could be intent, before and after, and $w_{<i}^{*}$ 
represents past tokens for each case.

\section{Experiments}
The goal of the experiments is to see how much we could reduce the annotated data and still achieve results
comparable to a case where the full human-annotated data is used. We tried different contextual data, which did not require
extra annotations, such as captions, facial expressions, and related concepts.

As shown in Fig.~\ref{fig:1}, we can intuitively see that some extra context could potentially help 
the language model to reach a more logical deduction of intention, past, and future events. 

For each scenario, the VisualCOMET dataset provides several human annotations for comparison,
each showing intent of a person, what could happen next, and what happened before.
The experiments are evaluated using BLEU \cite{papineni-etal-2002-bleu}, METEOR \cite{denkowski-lavie-2014-meteor} 
and CIDEr \cite{7299087} automatic metrics to compare the generated texts for different scenarios of 
before, intent and after with the human-annotated texts.

\begin{figure}
   \centering
   \begin{subfigure}[b]{0.4\textwidth}
       \centering
       \includegraphics[width=0.9\textwidth]{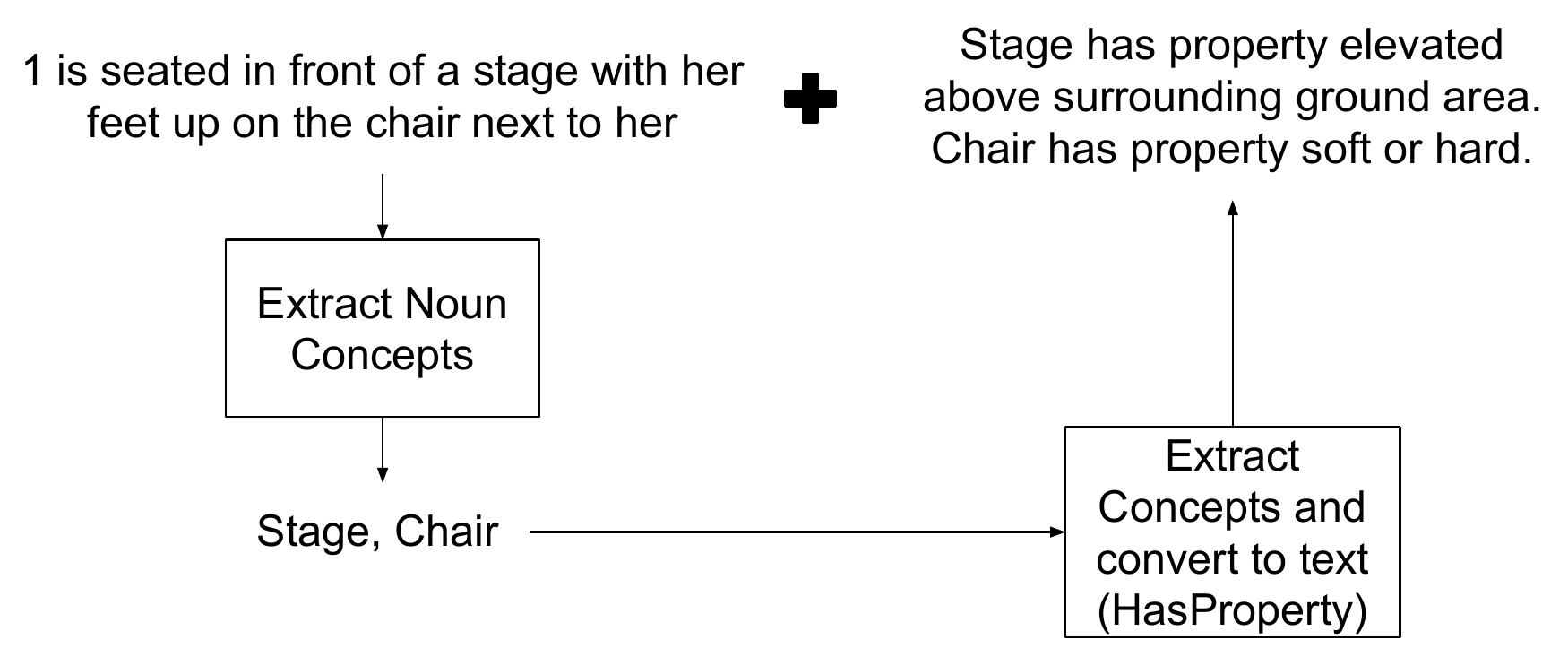}
       \caption{Process of adding relevant concepts.}
       \label{fig:2a}
   \end{subfigure}
   \hfill
   \begin{subfigure}[b]{0.4\textwidth}
       \centering
       \includegraphics[width=0.9\textwidth]{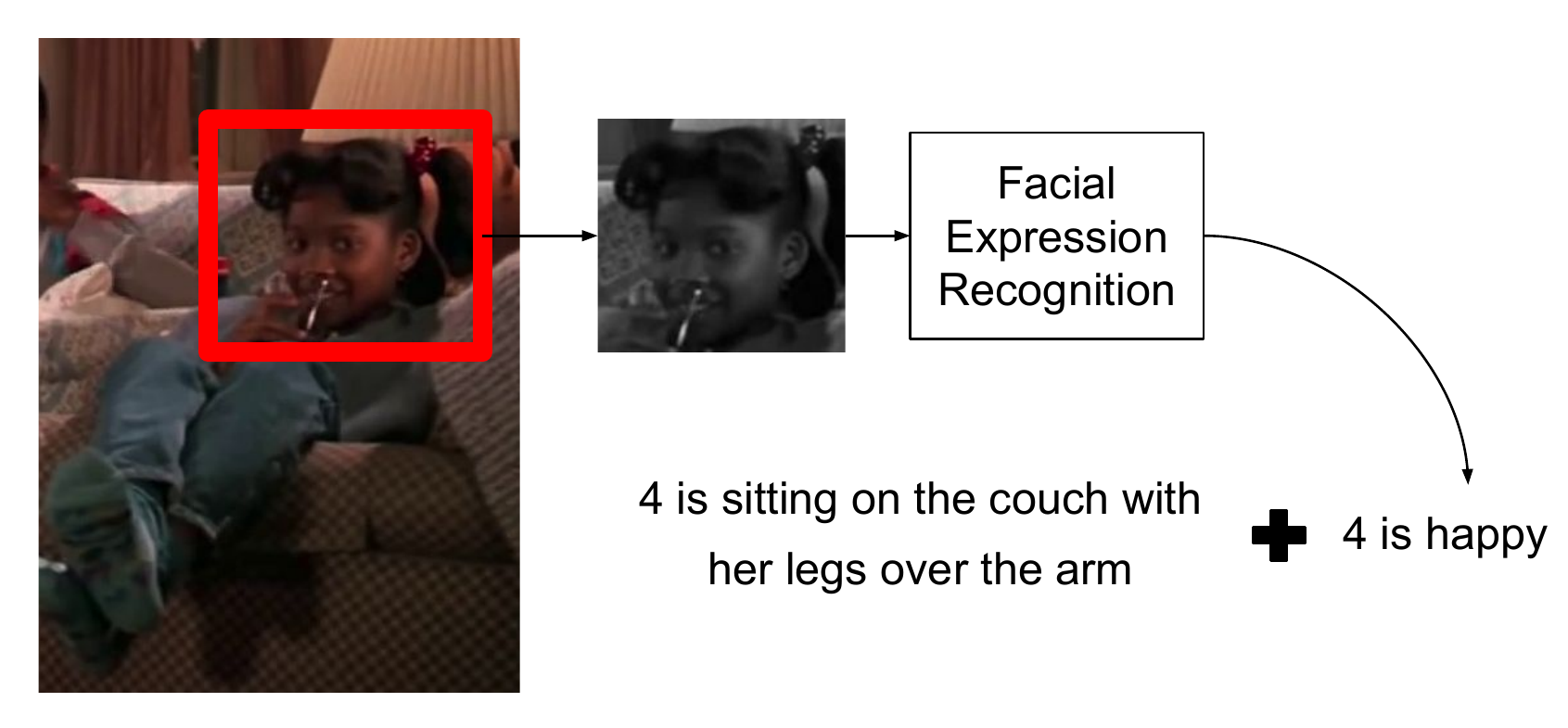}
       \caption{Process of adding facial expressions.}
       \label{fig:2b}
   \end{subfigure}
   \hfill
   \begin{subfigure}[b]{0.4\textwidth}
       \centering
       \includegraphics[width=0.9\textwidth]{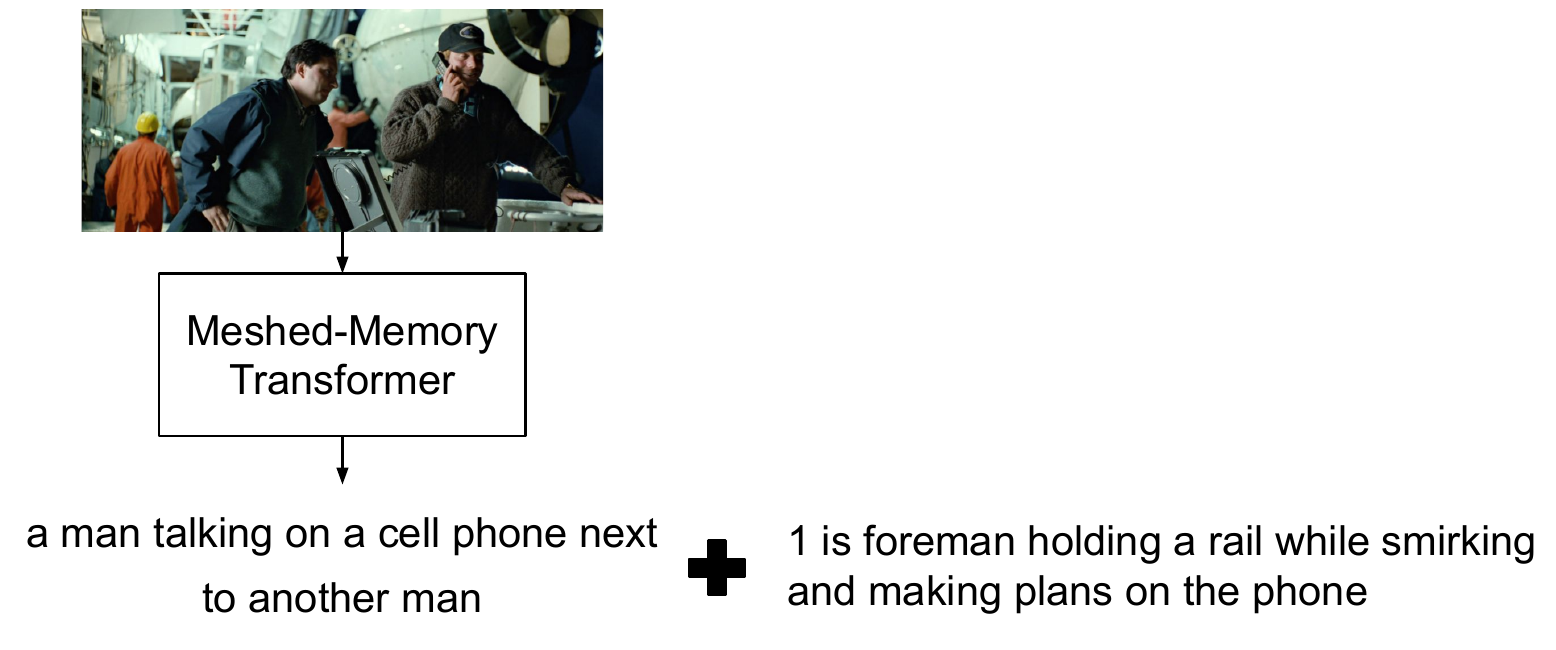}
       \caption{Process of adding image captions.}
       \label{fig:2c}
   \end{subfigure}
      \caption{The process of extracting and adding prompts shown through examples.}
      \label{fig:2}
\end{figure}

We tried two different methods of adding relevant concepts. One method is based on converting relevant concept graphs
into a readable sentence and the other method is based on only prepending concept words to the target sentence. 
In either method, the text is scanned for concepts, and the related concepts are extracted based on a commonsense knowledge graph 
such as in \cite{Speer2017ConceptNet5A, rezaei2020image}. Although sentence-based inputs
perform well, they require a longer input width that may not be available given the language model.
To sort relevant concepts, crowd-based scores or frequency scores are used based on the specific knowledge graph used. 
These triples are then converted to text with some hand-designed rules. 
An example of this process is shown in Fig.~\ref{fig:2a}. Table~\ref{tab:concept} 
shows three of the top-performing models with added conceptual contexts. They are compared
with the original data, which does not have any added context. Evaluation is done on a validation dataset 
with a size of a hundred. Concept words added in this specific scenario are connected via HasProperty and 
PartOf predicates. Concept sentences use the HasProperty predicate. Adding similar information during inference
time does not result in much improvement in this specific case. More comparisons can be found in the Appendix.

\begin{table}
   \begin{center}
   \begin{tabular}{|l|c|c|c|}
   \hline
   Method & BLEU-2 & METEOR & CIDEr \\
   \hline\hline
   GPT-2 \cite{Park2020VisualCOMETRA} & 13.81 & 10.85 & 15.37 \\
   Concept Word (NVP) & 17.25 & 12.17 & 19.79 \\
   Concept Word (VP) & 17.17 & 12.28 & 19.34 \\
   Concept Sent. (NVP) & 14.92 & 11.25 & 16.9 \\
   \hline
   \end{tabular}
   \end{center}
   \caption{Effects of adding relevant concepts. 
            Results are shown at the fourth epoch using almost 25,000 (22\%) 
            of the available annotated data. NVP: No Validation Prompt. 
            VP: Validation Prompt.}
   \label{tab:concept}
\end{table}

As seen in Fig.~\ref{fig:1a}, lack of the model's attention to some visual cues, such as facial expressions, 
could also result in errors of judgment. To fix this issue, we trained 
a ResNet \cite{He_2016} model on FER2013 \cite{Goodfellow2015ChallengesIR} dataset with 
almost $70\%$ accuracy. The dataset consists of human face images and emotion labels of 
angry, disgust, fear, happy, neutral, sad, and surprise. Only the emotion of people 
mentioned in the event text is processed. The results are then prepended to the event text. 
An example of this process is shown in Fig.~\ref{fig:2b}. Table \ref{tab:fe} shows effects of adding
facial expressions as a context in the final performance of the model. 
Contrary to relevant concepts, adding facial expressions during inference time 
improves the results.
Evaluation is done on a validation data size of a hundred.

\begin{table}
   \begin{center}
   \begin{tabular}{|l|c|c|c|}
   \hline
   Method & BLEU-2 & METEOR & CIDEr \\
   \hline\hline
   GPT-2 \cite{Park2020VisualCOMETRA} & 13.81 & 10.85 & 15.37 \\
   FE (NVP) & 14.45 & 11.27 & 15.7 \\
   FE (VP) & 15.11 & 11.23 & 16.03 \\
   \hline
   \end{tabular}
   \end{center}
   \caption{Effects of adding information about facial expressions.
            Results are shown at the fourth epoch using almost 25,000 (22\%) 
            of the available annotated data. NVP: No Validation Prompt. 
            VP: Validation Prompt. FE: Facial Expressions.}
   \label{tab:fe}
\end{table}

Another type of automatically-generated context that we experimented with is image captioning.
The idea is that some of the image dynamics may have been missed, even though
image features are fed into the GPT-2 model. Adding generated captions proves to be effective 
as shown in Table \ref{tab:captions}. Meshed-Memory transformer model \cite{cornia2020m2} 
with beam search decoding is used for image captioning. The process of adding these captions
is illustrated in Fig. \ref{fig:2c}.

\begin{table}
   \begin{center}
   \begin{tabular}{|l|c|c|c|}
   \hline
   Method & BLEU-2 & METEOR & CIDEr \\
   \hline\hline
   GPT-2 \cite{Park2020VisualCOMETRA} & 13.81 & 10.85 & 15.37 \\
   Caption (NVP) & 14.08 & 10.78 & 15.63 \\
   Caption (VP) & 16.49 & 11.85 & 18.8 \\
   \hline
   \end{tabular}
   \end{center}
   \caption{Effects of adding image captions. Results are shown at the fourth epoch using almost 25,000 (22\%) 
            of the available annotated data. NVP: No Validation Prompt. 
            VP: Validation Prompt.}
   \label{tab:captions}
\end{table}

A mixture of different contextual information is shown to be more effective than individual ones. 
A combination of concept words, image captions, and facial expressions of relevant 
individuals in the image achieve the best result compared to other experiments. 
As seen in Table \ref{tab:mixture}, this combination can achieve comparable results to 
full-data finetuning by only using ~35\%-40\% of the annotated data. 
This results in less human time spent doing annotations and can potentially 
reduce costs and completion times of projects. Results of other experiments are included in the Appendix.

\begin{table*}
   \begin{center}
   \begin{tabular}{|l|c|c|c|c|c|}
   \hline
   Method & Inference Data & Data Size & BLEU-2 & METEOR & CIDEr \\
   \hline\hline
   GPT-2 \cite{Park2020VisualCOMETRA} & N/A & 111,796 (100\%) & 18.05 & \textbf{13.21} & 22.72 \\
   CW + C + FE & C + CW + FE & 39,000 (~35\%) & 18.38 & 12.97 & 22.65 \\
   CW + C + FE & C + CW + FE & 45,000 (~40\%) & \textbf{18.58} & 13.01 & \textbf{22.97} \\
   \hline
   \end{tabular}
   \end{center}
   \caption{Analyzing the effect of combining multiple contextual data. 
            All models are finetuned for five epochs. 
            Contexts are added based on the order shown. 
            CW: Concept Words. C: Captions. FE: Facial Expressions.}
   \label{tab:mixture}
\end{table*}

To reduce the effects of other variables in these experiments, we have limited ourselves to 
only train the final models for five epochs. The decoding method and hyperparameters are
also kept constant throughout the experiments. We use nucleus sampling \cite{holtzman2019curious} 
with $p = 0.9$ to generate five sentences for each scenario of intent, before and after. 
The finetuning was run on two NVIDIA RTX GPUs with 24~GB memory each. 
For the case with all concept words, captions, and facial expression contexts, the fine-tuning
time is around 1.5 hours per epoch while using mixed precision.

\section{Conclusion}
In this work, we analyzed the effects of automatically-generated contexts 
in multimodal transformer models used in a commonsensical task. These prompts
can help us reduce the human annotation needed in the task by as much as 60\%-65\% 
and still, achieve comparable results to when the whole human-annotated dataset is used.
These findings result in time and cost savings for future multimodal
data annotation projects. 

As future work, it is interesting to find a lower bound for data annotation 
reduction without affecting the final result of a model. 
It is also useful to find a method to automatically 
find and apply the best contextual data for different tasks and models.

{\small
\bibliographystyle{ieee_fullname}
\bibliography{egbib}
}

\clearpage

\onecolumn
\appendix
\section{Appendix}

Experimentation results from using different types of training and inference prompts are included in this appendix. 
The model used in the experiments is GPT-2 as described in \cite{Park2020VisualCOMETRA}.


The best types of prompts are chosen to be combined. The experiments show that the order in which prompts are added can affect the final results.


The vision-based inference prompts seem to better affect the final metric results when compared to the text-based inference prompts. This could be 
due to the lack of enough visual attention paid during the decoding process. 
Future work could involve developing a multimodal model that makes better
use of visual contexts not only during the training phase, but also the inference time.

The quality of the annotated data can have an impact on the training model. We do not hand-select the annotated data based on quality and this may result in variability in final results when training with different data sizes. It can be a good practice to assess the quality of the annotated data and prompts based on the final goal of the model.

\begin{table*}[h!]
    \begin{center}
    \begin{tabular}{|c|c|c|c|c|c|}
    \hline
    Training Prompt & Inference Prompt & Training Data Size & BLEU-2 & METEOR & CIDEr \\
    \hline\hline
    None & None & 111,796 (100\%) & 17.94 & 13.14 & 22.71 \\
    None & None & 25,000 (~22\%) & 13.81 & 10.85 & 15.37 \\

    CS (AtLocation) & None & 25,000 (~22\%) & 12.26 & 10.42 & 15.09 \\
    CS (AtLocation) + Place & None & 25,000 (~22\%) & 14.3 & 11.08 & 14.66 \\

    CS (CapableOf) & None & 25,000 (~22\%) & 12.65 & 10.6 & 15.9 \\
    CS (CapableOf) + Place & None & 25,000 (~22\%) & 14.78 & 11.16 & 15.1 \\

    CS (HasA) & None & 25,000 (~22\%) & 12.73 & 10.65 & 15.83 \\
    CS (HasA) + Place & None & 25,000 (~22\%) & 14.58 & 11.15 & 15 \\

    CS (HasProperty) & None & 25,000 (~22\%) & 12.25 & 10.5 & 15.52 \\
    CS (HasProperty) + Place & None & 25,000 (~22\%) & 15.25 & 11.38 & 16.3 \\

    CS (IsA) & None & 25,000 (~22\%) & 12.73 & 10.41 & 15.73 \\
    CS (IsA) + Place & None & 25,000 (~22\%) & 14.04 & 11.03 & 14.32 \\

    CS (PartOf) & None & 25,000 (~22\%) & 12.65 & 10.51 & 15.71 \\
    CS (PartOf) + Place & None & 25,000 (~22\%) & 14.26 & 11.19 & 14.64 \\

    FE & None & 25,000 (~22\%) & 14.45 & 11.27 & 15.7 \\
    FE & FE & 25,000 (~22\%) & 15.11 & 11.23 & 16.03 \\

    CW (PartOf + HasProperty) & None & 25,000 (~22\%) & 17.25 & 12.17 & 19.79 \\
    CW (PartOf + HasProperty) & CW (PartOf + HasProperty) & 25,000 (~22\%) & 17.17 & 12.28 & 19.34 \\

    C & None & 25,000 (~22\%) & 14.08 & 10.78 & 15.63 \\
    C & C & 25,000 (~22\%) & 16.49 & 11.85 & 18.8 \\

    C + FE & C + FE & 25,000 (~22\%) & 16.75 & 12.19 & 19.16 \\

    CW + C + FE & None & 25,000 (~22\%) & 12.6 & 10.18 & 14.57 \\
    CW + C + FE & CW + C + FE & 25,000 (~22\%) & 17.4 & 11.97 & 20.03 \\
    
    CW + C + FE & CW + FE & 39,000 (~35\%) & 16.57 & 12.32 & 19.68 \\ 
    CW + C + FE & C + FE & 39,000 (~35\%) & 16.71 & 12.27 & 19.01 \\

    CW + C + FE & CW + C + FE & 39,000 (~35\%) & 16.75 & 12.4 & 19.86 \\
    CW + C + FE & CW + C + FE + Syns & 39,000 (~35\%) & 16.7 & 12.43 & 19.94 \\ 
    CW + C + FE + PCW & CW + C + FE + PCW & 39,000 (~35\%) & 17.74 & 12.73 & 20.52 \\
    CW + C + FE & C + CW + FE & 39,000 (~35\%) & 17.34 & 12.45 & 20.11 \\

    CW + C + FE & CW + C + FE & 45,000 (~40\%) & 17.46 & 12.84 & 21.56 \\
    CW + C + FE & C + CW + FE & 45,000 (~40\%) & 17.66 & 12.95 & 21.41 \\

    \hline
    \end{tabular}
    \end{center}
    \caption{Experimentation results of using different training and inference prompts. The model used is GPT-2 \cite{Park2020VisualCOMETRA}.
             Results are shown at epoch four and evaluated on validation data of size 100.
             Prompts are added based on the order shown. 
             CW: Concept Words. C: Captions. FE: Facial Expressions.
             CS: Concept Sentences. Syns: Synonyms. PCW: Place Concept Words.}
    \label{tab:experiments}
 \end{table*}

\end{document}